\documentclass[11pt]{article}

\usepackage[preprint]{acl}

\usepackage{times}
\usepackage{latexsym}

\usepackage[T1]{fontenc}

\usepackage[utf8]{inputenc}

\usepackage{microtype}

\usepackage{inconsolata}
\usepackage{amsmath}

\usepackage{graphicx}
\usepackage{tabularx}
\usepackage{booktabs}
\usepackage{multirow}
\usepackage{threeparttable}

\usepackage{ulem}

\usepackage{amsfonts}
\usepackage{bbm}

\title{Traj-Evolve: A Self-Evolving Multi-Agent System for Patient Trajectory Modeling in Lung Cancer Early Detection}

\author{
 \textbf{Sihang Zeng\textsuperscript{1,2}},
 \textbf{Matthew Thompson\textsuperscript{3}},
 \textbf{Ruth Etzioni\textsuperscript{2}},
 \textbf{Meliha Yetisgen\textsuperscript{1}}
 \\
 \textsuperscript{1}University of Washington,
 \textsuperscript{2}Fred Hutch Cancer Center,
 \textsuperscript{3}Google
 \\
 \texttt{\href{mailto:zengsh@uw.edu}{zengsh@uw.edu}}, \texttt{\href{mailto:melihay@uw.edu}{melihay@uw.edu}}
}

\begin{document}
\maketitle
\begin{abstract}
Modeling patient trajectories from longitudinal electronic health records (EHRs) requires reasoning over sparse, noisy, and long-context multimodal sequences. Existing LLM-based multi-agent systems address context length but process patients in isolation, failing to mirror how clinicians leverage accumulated experience from similar prior cases.
We present Traj-Evolve, a self-evolving multi-agent system with two complementary evolving mechanisms. First, an Experience Pool (ExPool) acts as a non-parametric memory, indexing rejection-sampled reasoning traces to retrieve similar patients as few-shot contexts. Second, multi-agent reinforcement learning (MARL) via reward-ranked fine-tuning parametrically optimizes inter-agent and agent-memory collaboration. A leave-one-out cross-retrieval strategy unifies the two, aligning training- and inference-time behavior under retrieval augmentation.
On a lung cancer prediction task utilizing up to five years of multimodal EHRs, Traj-Evolve outperforms 9 strong baselines on the overall population and a challenging never-smoker population. Analysis of the evolving dynamics highlights three key findings: (1) expanding the ExPool shifts optimal retrieval from diverse to specific samples; (2) under MARL, the manager agent's prediction loss converges quickly while the worker agents' temporal reasoning continues to benefit from more verified patients; and (3) the two mechanisms are complementary on the predicted risk, where ExPool improves specificity while MARL improves sensitivity.

\end{abstract}

\section{Introduction}

Lung cancer is the leading cause of cancer-related mortality worldwide \cite{sung2021global,lancaster2022low}, and early detection substantially improves patient outcomes \cite{lancaster2022low}. Longitudinal electronic health records (EHRs) offer a uniquely powerful opportunity for early detection, as they accumulate a rich, multimodal clinical history including diagnoses, procedures, laboratory values, vital signs, medications, and unstructured clinical notes, which collectively encode subtle disease trajectories preceding a cancer diagnosis \cite{jensen2012mining,kim2019ehr}. Within these trajectories lie early signals of risk and their trends, such as recurrent respiratory symptoms, chronic pulmonary conditions, or incidental radiographic findings documented years before diagnosis \cite{d2025inflammatory,ganti2021update}.

Extracting and temporally reasoning over these signals from long and noisy patient trajectories, however, is challenging. Recent studies evaluated LLM-based approaches for generalizable modeling from heterogeneous EHR data \cite{cui2025timer,kruse2025large,zeng2025trajcoa,zeng2026trajonco}. Among these, Traj-CoA is a multi-agent framework that leverages chain-of-agents and a long-term memory to facilitate temporal reasoning over patient trajectories for cancer early detection, eliminating complex feature engineering while achieving zero-shot performance comparable to supervised machine learning and deep learning models \cite{zeng2025trajcoa,zeng2026trajonco}.

Despite these advances, existing LLM-based longitudinal EHR modeling systems share a fundamental limitation: they are static. Every patient is processed in isolation, relying solely on the LLM's frozen parametric knowledge and a fixed prompt. This stands in sharp contrast to expert clinical practice, where diagnostic judgement is continually refined by accumulated experience with similar patients. This process is central to how clinicians recognise atypical presentations, such as early lung cancer in a never-smoker with an otherwise unremarkable history~\cite{eva2005casebased,patel2005thinking}. For lung cancer early detection, where cases are clinically heterogeneous and often subtly distinguished from controls by patterns distributed across years of records, the inability of a system to learn from past verified cases can limit both performance and robustness, particularly in minority subgroups such as never-smokers.
 
Emerging research on self-evolving LLM agents has the potential to address this gap. Rather than treating the model as immutable, self-evolving agents continually update their behavior through interaction and feedback, evolving their memory, prompts, tools, or parameters as new experience accumulates~\cite{gao2025selfevolving,zhang2025memorysurvey}. For example, memory-based approaches save the problem-solving trajectories as experience into an external database, which could guide future decisions through retrieval-augmented generation (RAG) \cite{shinn2023reflexion,zhao2024expel,wu2025evolver,zhou2025memento,tang2025agent}. 
In parallel, reinforcement-learning (RL) based approaches such as reward-ranked fine-tuning (RAFT)~\cite{dong2023raft,xiong2025minimalist, zhangSurveyReinforcementLearning2025} and multi-agent RL variants~\cite{ma2024coevolving,liao2025marft,zhang2025marti} enable collaborative agent systems to internalise successful reasoning patterns directly into their parameters. 

In healthcare, self-evolving agents have been explored in synthetic or simulated patient interactions \cite{li2024agenthospital,almansoori2025medagentsim} and medical question answering \cite{chen2025mdteamgpt}. Designing self-evolving systems for patient trajectory modeling in early cancer detection poses a distinct challenge: it requires complex temporal reasoning over years of noisy, multimodal data and the ability to draw reusable insights from heterogeneous clinical cases. Existing techniques may not be readily applicable to this scenario, and performance remains unclear. To our knowledge, no prior work has designed self-evolving agents to enhance longitudinal EHR modeling for real-world early cancer detection.
 
To bridge this gap, we present Traj-Evolve, a self-evolving multi-agent framework for patient trajectory modelling that extends the Traj-CoA architecture~\cite{zeng2025trajcoa} with two complementary evolutionary mechanisms, an evolving experience pool (ExPool) and multi-agent reinforcement learning (MARL). Collectively, these mechanisms enable Traj-Evolve to learn from its own experience as it processes more patients, continually refining its temporal reasoning and learning from ``patients-like-me", eventually improving performance over time. These two mechanisms transform patient trajectory modeling from a static, isolated prediction task into a continuously improving clinical learning system.
 
We evaluated Traj-Evolve on a large longitudinal cohort from a medical center, using five years of multimodal EHR history to predict incident lung cancer within the subsequent year among the overall population and the particularly challenging subgroup of never-smokers. We benchmarked against a comprehensive suite of baselines spanning clinical risk models, supervised machine learning, sequential deep learning, clinical BERT-based models, and LLM-based systems.
 
The main contributions of this work are:
\begin{itemize}
  \setlength\itemsep{0pt}
  \setlength\parskip{0pt}
  \item We introduce Traj-Evolve, to our knowledge, the first self-evolving multi-agent framework for longitudinal EHR modelling applied to a real-world clinical prediction task.
  \item We design two complementary evolutionary mechanisms: an evolving experience pool (ExPool) that provides non-parametric, few-shot ``patients-like-me'' retrieval, and a MARL procedure that parametrically optimizes inter-agent and agent-memory collaboration using rejection-sampled high-reward trajectories.
  \item We demonstrate that Traj-Evolve achieves state-of-the-art discrimination for one-year lung cancer prediction in the overall population and the challenging never-smoker population.
  \item We provide detailed analyses of the self-evolving dynamics, supporting the vision of a continuously improving clinical decision-support system.
\end{itemize}

\begin{figure}[h!]
    \centering
    \includegraphics[width=\linewidth]{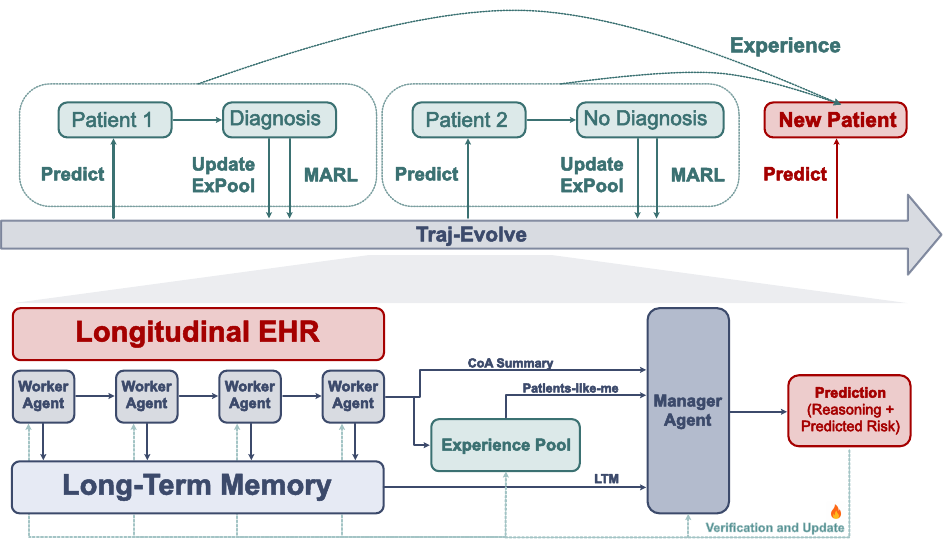}
    \caption{\textbf{Overview of the Traj-Evolve architecture and self-evolving workflow.} The top panel illustrates the self-evolving process, wherein the system accumulates experience from prior verified patients to iteratively update Traj-Evolve and facilitate prediction for a new patient. The bottom panel details the pipeline.}
    \label{fig:architecture}
\end{figure}

\section{Related Works}
\paragraph{LLM-based Patient Trajectory Modeling}
Recent work increasingly leverages strong LLMs for zero- or few-shot reasoning over heterogeneous clinical histories, including DT-GPT for clinical variable forecasting \cite{makarov2025large}, EHR2Path for scalable patient pathway prediction \cite{pellegrini2025ehr2path}, TIMER for temporal instruction tuning \cite{cui2025timer}, and \citet{kruse2025large} for long-context summarization. Yet single-LLM pipelines remain limited by the lost-in-the-middle phenomenon \cite{liuLostMiddleHow2024} on very long EHRs and by the complexity in specific clinical prediction tasks, motivating multi-agent designs to decompose longitudinal EHR modeling into simpler subtasks. MoMA \cite{gaoMoMAMixtureofmultimodalagentsArchitecture2025} coordinates modality-specialized agents for clinical prediction, CARE-AD \cite{liCAREADMultiagentLarge2025a} and ClinNoteAgents \cite{zhouClinNoteAgentsLLMMultiAgent2025} decompose reasoning across specialist agents, and Traj-CoA \cite{zeng2025trajcoa, zeng2026trajonco} extend chain-of-agents \cite{zhangChainAgentsLarge2024} with long-term memory for cancer early detection. Complementary efforts such as CliCARE \cite{liCliCAREGroundingLarge2025} and TRACE \cite{quTRACETemporalReasoning2026} further explore temporal knowledge graph and dual-memory approaches. However, these systems are static: each patient is reasoned about in isolation, with no mechanism or evaluation for accumulating verified clinical experience over time.

\paragraph{Self-Evolving Agents}
\citet{gao2025selfevolving} organize self-evolving agents along three axes: what, when, and how to evolve. Along what to evolve, prior work targets memory, prompts, tools, or model parameters; along when, adaptation can be intra- or inter-test-time; along how, it is driven by textual feedback or scalar rewards in single- or multi-agent settings. For example, memory-evolving methods such as Reflexion \cite{shinn2023reflexion}, ExpeL \cite{zhao2024expel}, Memento \cite{zhou2025memento}, and Agent KB \cite{tang2025agent} store and retrieve past trajectories as non-parametric experience. Parameter-evolving methods internalize successful experience via model training, including supervised fine-tuning and reinforcement learning \cite{zelikmanSTaRBootstrappingReasoning2022, zuoTTRLTestTimeReinforcement2025, dong2023raft, wangRAGENUnderstandingSelfEvolution2025} and multi-agent extensions \cite{zhang2025marti, ma2024coevolving, liao2025marft}. These two families are typically pursued in isolation. In healthcare, self-evolution has so far been confined to simulated or interactive settings, including Agent Hospital \cite{li2024agenthospital}, MedAgentSim \cite{almansoori2025medagentsim}, MDTeamGPT \cite{chen2025mdteamgpt}, and EvoClinician \cite{heEvoClinicianSelfEvolvingAgent2026}. 

To our knowledge, no prior work has applied self-evolving agents to longitudinal EHR modeling for real-world clinical prediction. Traj-Evolve fills this gap by jointly evolving memory (ExPool) and parameters (MARL) at inter-test-time, unified by a leave-one-out cross-retrieval procedure that aligns training- and inference-time augmentation for lung cancer early detection.

\section{Methods}
\label{sec:method}
 
\subsection{Problem Formulation}
\label{sec:problem}
 
\paragraph{Lung Cancer Early Detection}
Let $\mathcal{P} = \{p_i\}_{i=1}^{N}$ denote a cohort of patients. For each patient $p_i$, we observe a longitudinal multimodal EHR sequence
\begin{equation}
\mathcal{X}_i = \big\{(t_{i,j}, e_{i,j})\big\}_{j=1}^{T_i},
\quad t_{i,j} \le t_i^{\star},
\end{equation}
where $t_i^{\star}$ is the patient-specific index date (time of prediction), $T_i$ is the number of dated entries within the available EHR, and each event $e_{i,j}$ at time $t_{i,j}$ consists of either a structured record (diagnosis, medication, lab, vital, or procedure code) or unstructured clinical text (notes and radiology reports). The binary target $y_i \in \{0,1\}$ indicates whether $p_i$ receives a first primary lung-cancer diagnosis within one year after $t_i^{\star}$.
 
The task of lung cancer early detection is to learn a function $f_\theta: \mathcal{X}_i \mapsto (s_i, r_i)$ that maps the longitudinal record to an integer risk score $s_i \in \{1,\dots,10\}$ and a natural-language rationale $r_i$. 

\paragraph{Self-Evolving System}
Beyond standard generalization, we additionally require $f_\theta$ to improve as more patients are seen and verified, yielding a self-evolving system. We formalize this as follows. Patients arrive sequentially as a stream $p_1, p_2, \dots$. At each step $t$, the system maintains an experience set $\mathcal{E}_t$ that summarises all patients that have been processed and verified by a known diagnostic status so far. The primary objective of a self-evolving system is to continuously improve the performance of $f_{\theta(\mathcal{E}_t)}$ as the experience set $\mathcal{E}_t$ grows. This captures two coupled challenges: $f_{\theta(\mathcal{E}_t)}$ must perform complex temporal reasoning over long and noisy EHRs to produce $(s_t, r_t)$, and it additionally needs to construct and leverage $\mathcal{E}_t$ so that performance improves with $|\mathcal{E}_t|$.

We randomly split the dataset into training $\mathcal{D}_{tr}$, validation $\mathcal{D}_{val}$, and test set $\mathcal{D}_{test}$. In practice, we simulated the stream of verified patients by drawing a random sample from the training set that grew incrementally over time.

\subsection{Background: Traj-CoA Base System}
\label{sec:base}
 
We build on Traj-CoA \citep{zeng2025trajcoa, zeng2026trajonco}, a static chain-of-agents (CoA) \cite{zhangChainAgentsLarge2024} backbone that handles long-context EHRs. Each $\mathcal{X}_i$ is serialized into an LLM-friendly XML representation and segmented into $C_i$ chronologically ordered chunks $\{c_{i,1},\dots,c_{i,C_i}\}$ such that each chunk fits within the LLM context limit.
 
A sequence of $C_i$ worker agents, all parameterized by $\theta_w$, processes the chunks sequentially. Concretely, the $\ell$-th worker maintains a running summary $u_{i,\ell}$ and a long-term episodic memory (LTM) $\mathcal{M}_{i,\ell}$. The final worker summary $u_i=u_{i,C_i}$ encapsulates the full EHR trajectory, while $\mathcal{M}_i=\mathcal{M}_{i,C_i}$ stores a condensed timeline of lung-cancer-related events extracted across all chunks. A manager agent $\pi_{\theta_m}$ then synthesizes both signals to produce the final output:
\begin{equation}
(s_i, r_i) = \pi_{\theta_m}\big(u_i, \mathcal{M}_i\big).
\end{equation}
 
This zero-shot pipeline decomposes complex temporal reasoning into inter-agent and agent-memory collaboration tasks, improving patient trajectory modeling performance. However, it treats each patient in isolation, conditioning only on the LLM's parametric knowledge. This creates a gap with expert clinical practice, which relies on accumulated case experience. To bridge this gap, we design two complementary self-evolving mechanisms atop Traj-CoA for our self-evolving Traj-Evolve system (Figure \ref{fig:architecture}).

\begin{figure}[htbp!]
    \centering
    \includegraphics[width=\linewidth]{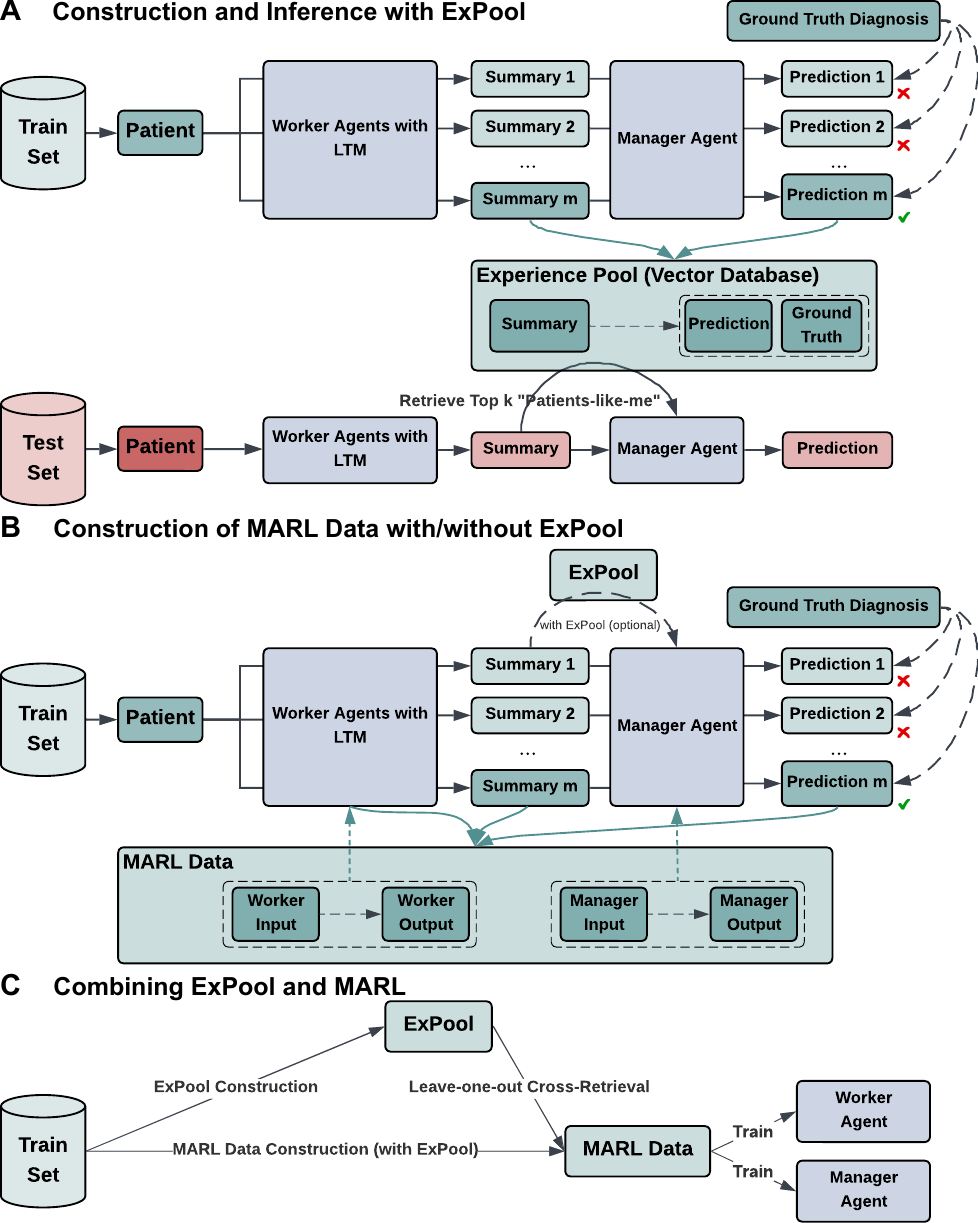}
    \caption{\textbf{Methodological design of the self-evolving mechanisms.} (A) Construction and inference pipeline for the ExPool. (B) Construction of MARL training data. (C) Integration of ExPool and MARL.}
    \label{fig:method}
\end{figure}

\subsection{Evolving Experience Pool (ExPool)}
\label{sec:expool}
 
Inspired by the procedural memory approach that saves successful reasoning traces for future problem solving of similar tasks \cite{tang2025agent}, we design an evolving experience pool (ExPool, Figure \ref{fig:method}A). ExPool equips Traj-Evolve with a non-parametric procedural memory that saves certain reasoning traces of verified patients. As Traj-Evolve generates predictions that can be subsequently verified against ground-truth diagnostic status (e.g., confirmed cancer diagnosis or benign status), these verified reasoning traces can serve as experience for future cases. This design is analogous to expert clinical reasoning, in which a patient's presentation is rarely adjudicated in isolation but rather by analogy to remembered cases that share a similar longitudinal clinical pattern. Correct reasoning traces may offer reusable diagnostic patterns, while incorrect predictions may expose model vulnerabilities and trigger self-reflection. These data-driven patterns evolve the system beyond its current boundary, forming the self-evolving capability.

\paragraph{ExPool Construction}
To capture the current system's boundary, i.e., the optimal diagnostic ability, we constructed ExPool using a rejection sampling approach that selected best-of-N from a set of roll-out reasoning traces. Formally, for each patient $p_i\in \mathcal{D}_{tr}$, we draw $m$ independent roll-outs from the base system under an elevated sampling temperature $\tau>1$ to diversify reasoning traces,
\begin{equation}
\big\{(u_i^{(j)}, \mathcal{M}_i^{(j)}, s_i^{(j)}, r_i^{(j)})\big\}_{j=1}^{m} \sim f_{\theta}(\mathcal{X}_i;\tau),
\end{equation}
and retain a single optimal trace via label-conditioned rejection sampling
\begin{equation}
j^{\star}_i =
\begin{cases}
\arg\max_{j} s_i^{(j)}, & y_i = 1, \\
\arg\min_{j} s_i^{(j)}, & y_i = 0.
\end{cases}
\label{eq:rejsamp}
\end{equation}
Notably, these optimal reasoning traces may mix correct and incorrect predictions, which provides different signals for ExPool. 

These selected traces then populate ExPool, a vector database where the retrieval keys are the embeddings of the final worker agent summaries: \begin{equation}
\mathbf{v}_i = \phi(u_i^{\star}),
\end{equation}
in which $\phi(\cdot)$ denotes an embedding model and index each experience as a key-value pair
\begin{equation}
\mathcal{E} = \Big\{\big(\mathbf{v}_i,\; (r_i^{\star}, s_i^{\star}, y_i)\big)\Big\}_{i \in \mathcal{D}_\text{tr}},
\end{equation}
where the value stores the manager's rationale, predicted risk, and ground-truth label. This design makes it feasible to embed and index long patient trajectories in the latent space, providing an efficient approach for experience retrieval.

\paragraph{ExPool Inference}
During inference, ExPool functions using a retrieval-augmented generation (RAG) \cite{xuRetrievalAugmentedGenerationKnowledge2024} mechanism. For each new patient $p_q$, we adopt semantic retrieval by querying $\mathcal{E}$ with $\mathbf{v}_q = \phi(u_q)$ and returning the top-$k$ nearest neighbors as ``patients-like-me'' using cosine similarity:
\begin{equation}
\mathcal{N}_k(q) = [\operatorname*{Top-} k]_{i \in \mathcal{E}} \;\frac{\mathbf{v}_q^{\top}\mathbf{v}_i}{\|\mathbf{v}_q\|\,\|\mathbf{v}_i\|}.
\end{equation}
The semantic matching is chosen over exact matching because it yields a soft neighborhood that balances diversity and specificity. Retrieved patients may exhibit different clinical profiles, matching the patient $p_q$ in diverse ways. Rather than collapsing this heterogeneity into a hard label via majority voting, we delegate the comparative reasoning to the manager agent, conditioning it on the full retrieved set:
\begin{equation}
(s_q, r_q) = \pi_{\theta_m}\big(u_q,\, \mathcal{M}_q,\, \{(r_i^{\star}, s_i^{\star}, y_i)\}_{i\in\mathcal{N}_k(q)}\big).
\end{equation}

ExPool therefore provides the manager agent with diverse but clinically relevant ``patients-like-me'' as in-context examples, and the manager agent comprehensively reasons over all information for prediction. As ExPool continuously scales with newly verified patients, the retrieval of highly specific neighbors becomes increasingly precise, shifting the system from isolated predictions to a progressively self-evolving framework.

\subsection{Multi-Agent Reinforcement Learning (MARL)}
\label{sec:marl}
 
ExPool adapts the system at inference time, but it leaves the underlying agent parameters $(\theta_w, \theta_m)$ unchanged and acts only on the final manager step. The intermediate worker reasoning, which decides what gets distilled into $\mathcal{M}_i$ (agent-memory communication) and how the chunk-by-chunk summary is built up (inter-agent communication), does not receive a learning signal from the verified outcomes. To close this gap, we introduce a second self-evolving mechanism that internalizes successful reasoning traces directly into the model parameters, adapting reward-ranked fine-tuning (RAFT) \citep{dong2023raft}, a reinforcement learning strategy that optimizes models using the reasoning trace with the highest reward among all self-generated roll-outs, to our multi-agent setting. (Figure \ref{fig:method}B)

\paragraph{Reward and Accepted Set}
Parallel to the construction of ExPool, we generated $m$ roll-outs with an elevated temperature. We use the ground-truth label $y_i$ as a binary reward signal. Unlike ExPool, which retains the optimal trace per patient regardless of its correctness, MARL retains only clinically consistent traces in its accepted set $\mathcal{A}$:
\begin{equation}
\mathcal{A} = \big\{(i,j) : R(s_i^{(j)}, y_i) = 1\big\},
\end{equation}
where the reward function is
\begin{equation}
R(s,y) =
\begin{cases}
\mathbbm{1}[s \ge 6], & y = 1, \\
\mathbbm{1}[s \le 4], & y = 0.
\end{cases}
\label{eq:reward}
\end{equation}
Eq.~\eqref{eq:reward} acts as a hard rejection filter, ensuring the resulting fine-tuning data consists exclusively of logically sound and clinically accurate reasoning traces.
 
\paragraph{Decoupled Optimization}
For each accepted trace $(i,j)\in\mathcal{A}$, we decompose the roll-out into per-agent input-output pairs. Let $\mathbf{x}^{w}_{i,j,c}$ and $\mathbf{y}^{w}_{i,j,c}$ denote the input and output of the $c$-th worker, and let $\mathbf{x}^{m}_{i,j}, \mathbf{y}^{m}_{i,j}$ denote the manager pair. To preserve temporal balance across the chain, we subsample worker positions to retain the first, the last, and two randomly selected intermediate worker agents per reasoning trace.
\begin{equation}
\mathcal{C}_{i,j} = \{1,\, C_i\} \,\cup\, \mathrm{Sample}_2\big(\{2,\dots,C_i-1\}\big),
\end{equation}
yielding the worker and manager training sets
\begin{align}
\mathcal{D}_w &= \big\{(\mathbf{x}^{w}_{i,j,c}, \mathbf{y}^{w}_{i,j,c}) : (i,j)\in\mathcal{A},\, c\in\mathcal{C}_{i,j}\big\}, \\
\mathcal{D}_m &= \big\{(\mathbf{x}^{m}_{i,j}, \mathbf{y}^{m}_{i,j}) : (i,j)\in\mathcal{A}\big\}.
\end{align}
 
The worker and manager parameters are updated independently to preserve their specialized roles:
\begin{align}
\theta_w^{\star} &= \arg\min_{\theta_w} \!\!\!\sum_{(\mathbf{x},\mathbf{y})\in\mathcal{D}_w}\!\!\! -\log \pi_{\theta_w}(\mathbf{y}\mid\mathbf{x}), \\
\theta_m^{\star} &= \arg\min_{\theta_m} \!\!\!\sum_{(\mathbf{x},\mathbf{y})\in\mathcal{D}_m}\!\!\! -\log \pi_{\theta_m}(\mathbf{y}\mid\mathbf{x}).
\end{align}
Optimizing $\theta_w$ refines sequential inter-agent reasoning and worker-memory collaboration, while optimizing $\theta_m$ refines how the final summary and the LTM are aggregated into a risk estimation. As verified patient cohort expands, this MARL approach benefits from larger training data and drives continuous self-evolution through reinforcing successful temporal reasoning and collaborative memory distillation to progressively enhance Traj-Evolve's intrinsic diagnostic capabilities.
 
\subsection{Combining ExPool and MARL}
\label{sec:combine}
 
ExPool provides retrieved ``patients-like-me'' but does not update parameters; Vanilla MARL updates parameters but lacks an explicit mechanism to incorporate similar patients during final risk estimation. We unify the two through a leave-one-out cross-retrieval procedure that prevents data leakage and matches training and inference input formats, while injecting retrieval signals into the optimization. (Figure \ref{fig:method}C)
 
For each $p_i \in \mathcal{D}_\text{tr}$, we construct a patient-specific dynamic pool $\mathcal{E}_{-i} = \mathcal{E} \setminus \{i\}$ from which we retrieve $\mathcal{N}_k(i)$ for the manager agent during the MARL roll-out phase. The manager input is augmented with the retrieved patients for prediction, after which the rejection filter in Eq.~\eqref{eq:reward} is applied to obtain the accepted set $\mathcal{A}^{\,\text{loo}}$. Formally, the accepted set is defined as:
\begin{equation}
\mathcal{A}^{\,\text{loo}} = \Big\{(i,j) : R\big(s_i^{(j)}(\mathcal{E}_{-i}),\, y_i\big) = 1\Big\},
\end{equation}
where $s_i^{(j)}(\mathcal{E}_{-i})$ denotes the $j$-th roll-out conditioned on neighbours retrieved from $\mathcal{E}_{-i}$. This methodological synthesis yields optimal traces that benefit from the augmented context provided by ``patients-like-me'' from the dynamic ExPool, thereby generating potentially better training data compared to MARL in isolation. The worker and manager agents were subsequently trained using the standard MARL protocol, while inference was conducted utilizing the standard ExPool RAG methodology on the full ExPool from all training patients.

\section{Experiments}
\label{sec:experiments}

\paragraph{Dataset}
We predict first primary lung cancer diagnoses within a one-year window using an in-house longitudinal EHR dataset. The input $\mathcal{X}_i$ comprises up to five years of EHR history before the index date $t_i^\star$ (the completion of a chest radiology exam). Cases and controls are 1:10 matched on index exam type and date. The training set $\mathcal{D}_{tr}$ contains $13,629$ patients. We evaluate on two disjoint, held-out test sets: an overall cohort ($n=1,000$) and a clinically more challenging never-smoker cohort ($n=835$). Notably, patients inputs are exceptionally long, with median XML token counts exceeding 60k overall and 80k for never-smokers (Table \ref{tab:baseline}; Appendix \ref{app:dataset}).

\paragraph{Implementation Details}
Traj-Evolve uses GPT-OSS-20B \cite{openaiGptoss120bGptoss20bModel2025} as the base LLM and nomic-embed-text-v1.5 \cite{nussbaumNomicEmbedTraining2024} for ExPool embeddings. During roll-out, we sample $m=4$ traces at temperature $\tau=1.5$. Both MARL agents are trained via QLoRA \cite{dettmersQLoRAEfficientFinetuning2023} for one epoch.

\paragraph{Baselines}
We compare against five baseline categories (Appendix \ref{app:baselines}): (1) clinical risk models (LCRAT \citep{katki2016development}); (2) supervised ML (Logistic Regression, XGBoost \citep{chenXGBoostScalableTree2016}); (3) sequential deep learning (RETAIN \citep{choi2016retain}, PatientTM \citep{silvaModellingPatientTrajectories2022a}); (4) clinical BERT (Clinical ModernBERT \citep{leeClinicalModernBERTEfficient2025}); and (5) GPT-OSS-20B-based LLM pipelines (vanilla LLM, RAG, and Traj-CoA \citep{zeng2025trajcoa}).
Further details are in Appendix \ref{app:baselines}.

\paragraph{Evaluation}
For binary risk classification, we report AUROC, AUPRC, and F1 as primary metrics, alongside sensitivity, specificity, PPV, and NPV from the predicted risk $s_i$ and binary target $y_i$. All results are reported as bootstrap means and standard errors over $1,000$ resamples of each test set.

\section{Results}
\label{sec:results}
\begin{table*}[htbp!]
\centering
\caption{Model performance comparison on overall population.}
\label{tab:1_year_performance}
\resizebox{\textwidth}{!}{
\begin{tabular}{llll ccccccc}
\toprule
\textbf{Model Family} & \textbf{Model} & \textbf{Fine-tuning} & \textbf{Data Modalities} & \textbf{AUROC} & \textbf{AUPRC} & \textbf{Sensitivity} & \textbf{Specificity} & \textbf{PPV} & \textbf{NPV} & \textbf{F1} \\
\midrule
Clinical & LCRAT & Zero-shot & Clinical & 0.690 (0.027) & 0.169 (0.028) & 0.789 (0.137) & 0.538 (0.140) & 0.154 (0.033) & 0.965 (0.014) & 0.252 (0.034) \\
\midrule
\multirow{2}{*}{ML} & LR & SFT & Codes & 0.752 (0.027) & 0.267 (0.042) & 0.705 (0.081) & 0.703 (0.074) & 0.197 (0.034) & 0.960 (0.009) & 0.304 (0.036) \\
& Xgboost & SFT & Codes & 0.755 (0.027) & 0.261 (0.042) & 0.713 (0.069) & 0.712 (0.057) & 0.201 (0.030) & 0.962 (0.008) & 0.312 (0.034) \\
\midrule
\multirow{2}{*}{DL} & RETAIN & SFT & Codes & 0.699 (0.029) & 0.195 (0.032) & 0.661 (0.158) & 0.654 (0.157) & 0.179 (0.056) & 0.953 (0.014) & 0.268 (0.044) \\
& PatientTM & SFT & Codes, Texts & 0.683 (0.028) & 0.189 (0.033) & 0.710 (0.105) & 0.595 (0.105) & 0.154 (0.026) & 0.955 (0.011) & 0.249 (0.029) \\
\midrule
BERT & Clinical ModernBERT & SFT & All & 0.740 (0.027) & 0.241 (0.040) & 0.674 (0.086) & 0.718 (0.077) & 0.199 (0.034) & 0.957 (0.009) & 0.303 (0.036) \\
\midrule
\multirow{6}{*}{LLM} & GPT-OSS-20B & Zero-shot & All & 0.794 (0.024) & 0.262 (0.040) & 0.734 (0.067) & 0.748 (0.062) & 0.230 (0.038) & 0.966 (0.008) & 0.347 (0.041) \\
& RAG & Zero-shot & All & 0.806 (0.024) & 0.296 (0.047) & 0.788 (0.067) & 0.712 (0.061) & 0.218 (0.036) & 0.972 (0.008) & 0.339 (0.040) \\
& Traj-CoA & Zero-shot & All & 0.814 (0.023) & 0.302 (0.059) & 0.727 (0.093) & 0.767 (0.099) & 0.253 (0.053) & 0.970 (0.010) & 0.371 (0.055) \\
& Traj-Evolve (ExPool) & Few-shot & All & 0.837 (0.020) & 0.315 (0.042) & 0.705 (0.049) & 0.805 (0.013) & 0.271 (0.030) & 0.964 (0.007) & 0.391 (0.035) \\
& Traj-Evolve (MARL) & SFT & All & 0.843 (0.023) & 0.309 (0.042) & \textbf{0.813} (0.045) & 0.798 (0.021) & 0.272 (0.032) & \textbf{0.979} (0.006) & 0.408 (0.037) \\
& Traj-Evolve (ExPool+MARL) & SFT & All & \textbf{0.860} (0.020) & \textbf{0.323} (0.045) & 0.724 (0.086) & \textbf{0.841} (0.094) & \textbf{0.293} (0.057) & 0.971 (0.009) & \textbf{0.417} (0.062) \\
\bottomrule
\end{tabular}
}
\end{table*}

\subsection{Main Results}
\paragraph{Overall Population} 
As shown in Table \ref{tab:1_year_performance}, Traj-Evolve achieves state-of-the-art discrimination in the overall population. The combined Traj-Evolve (ExPool+MARL) achieves the best AUROC (0.86), AUPRC (0.32), and F1 (0.42). It outperforms the strongest static LLM baseline, Traj-CoA (AUROC 0.81), as well as LCRAT (0.69), XGBoost (0.76), zero-shot GPT-OSS-20B (0.79), and RAG (0.81). The two self-evolving variants, ExPool (0.84) and MARL (0.84), also outperform all baselines when evaluated independently. These results show that both mechanisms contribute meaningful improvements.

\begin{table*}[htbp!]
\centering
\caption{Model performance comparison on never-smoker population}
\label{tab:1_year_performance_nonsmoker}
\resizebox{\textwidth}{!}{
\begin{tabular}{llll ccccccc}
\toprule
\textbf{Model Family} & \textbf{Model} & \textbf{Fine-tuning} & \textbf{Data Modalities} & \textbf{AUROC} & \textbf{AUPRC} & \textbf{Sensitivity} & \textbf{Specificity} & \textbf{PPV} & \textbf{NPV} & \textbf{F1} \\
\midrule
Clinical & LCRAT & Zero-shot & Clinical & 0.606 (0.046) & 0.046 (0.012) & 0.726 (0.123) & 0.580 (0.108) & 0.059 (0.024) & 0.985 (0.005) & 0.107 (0.031) \\
\midrule
\multirow{2}{*}{ML} & LR & SFT & Codes & 0.678 (0.046) & 0.055 (0.013) & 0.690 (0.103) & 0.685 (0.088) & 0.070 (0.017) & 0.985 (0.005) & 0.127 (0.027) \\
& Xgboost & SFT & Codes & 0.706 (0.050) & 0.078 (0.022) & 0.616 (0.092) & 0.819 (0.049) & 0.105 (0.026) & 0.985 (0.004) & 0.178 (0.038) \\
\midrule
\multirow{2}{*}{DL} & RETAIN & SFT & Codes & 0.571 (0.047) & 0.045 (0.015) & 0.776 (0.160) & 0.442 (0.157) & 0.047 (0.018) & 0.985 (0.007) & 0.087 (0.020) \\
& PatientTM & SFT & Codes, Texts & 0.623 (0.040) & 0.043 (0.009) & \textbf{0.834} (0.109) & 0.468 (0.117) & 0.051 (0.010) & 0.989 (0.006) & 0.096 (0.018) \\
\midrule
BERT & Clinical ModernBERT & SFT & All & 0.755 (0.036) & 0.103 (0.037) & 0.825 (0.087) & 0.640 (0.091) & 0.074 (0.017) & \textbf{0.991} (0.004) & 0.134 (0.028) \\
\midrule
\multirow{6}{*}{LLM} & GPT-OSS-20B & Zero-shot & All & 0.755 (0.049) & 0.197 (0.065) & 0.622 (0.105) & 0.823 (0.069) & 0.115 (0.044) & 0.985 (0.004) & 0.190 (0.065) \\
& RAG & Zero-shot & All & 0.780 (0.047) & 0.185 (0.057) & 0.697 (0.125) & 0.786 (0.111) & 0.120 (0.054) & 0.988 (0.005) & 0.196 (0.069) \\
& Traj-CoA & Zero-shot & All & 0.775 (0.048) & 0.227 (0.081) & 0.665 (0.127) & 0.817 (0.115) & 0.136 (0.061) & 0.987 (0.005) & 0.215 (0.080) \\
& Traj-Evolve (ExPool) & Few-shot & All & 0.819 (0.047) & 0.269 (0.087) & 0.542 (0.105) & \textbf{0.927} (0.009) & 0.194 (0.050) & 0.984 (0.005) & 0.286 (0.064) \\
& Traj-Evolve (MARL) & SFT & All & 0.822 (0.040) & 0.259 (0.062) & 0.728 (0.084) & 0.854 (0.049) & 0.158 (0.043) & 0.989 (0.004) & 0.257 (0.057) \\
& Traj-Evolve (ExPool+MARL) & SFT & All & \textbf{0.835} (0.048) & \textbf{0.279} (0.082) & 0.680 (0.096) & 0.905 (0.011) & \textbf{0.200} (0.048) & 0.988 (0.004) & \textbf{0.309} (0.057) \\
\bottomrule
\end{tabular}
}
\end{table*}

\paragraph{Never-smoker Population} 
Table \ref{tab:1_year_performance_nonsmoker} presents the never-smoker population results. For this cohort, traditional models degrade sharply (LCRAT 0.61; XGBoost 0.71), reflecting their reliance on smoking-related features. Traj-Evolve remains robust with the ExPool variant (0.82), MARL variant (0.82), and the combined system (AUROC 0.84; AUPRC 0.28; F1 0.31). These results demonstrate that the self-evolving design of Traj-Evolve can effectively adapt to this clinically challenging population.

\begin{figure}[h!]
    \centering
    \includegraphics[width=.75\linewidth]{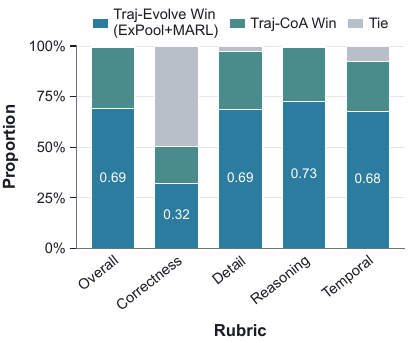}
    \caption{LLM-as-a-judge evaluation compares Traj-Evolve and Traj-CoA.}
    \vspace{-10pt}
    \label{fig:llm_judge}
\end{figure}

\paragraph{Reasoning Quality} 
Using a pairwise LLM-as-a-judge protocol similar to \citet{zeng2026trajonco}, we compare the final outputs from the static Traj-CoA and our Traj-Evolve system (Figure \ref{fig:llm_judge}). Traj-Evolve is preferred over Traj-CoA in 69\% of overall judgments, with consistent wins on detail, clinical reasoning, and temporal coherence rubrics (68\%--73\%). This indicates Traj-Evolve's reasoning quality also improves alongside discrimination.

\subsection{Evolutionary Dynamics of ExPool}
\begin{figure}[h!]
\centering
\includegraphics[width=\linewidth]{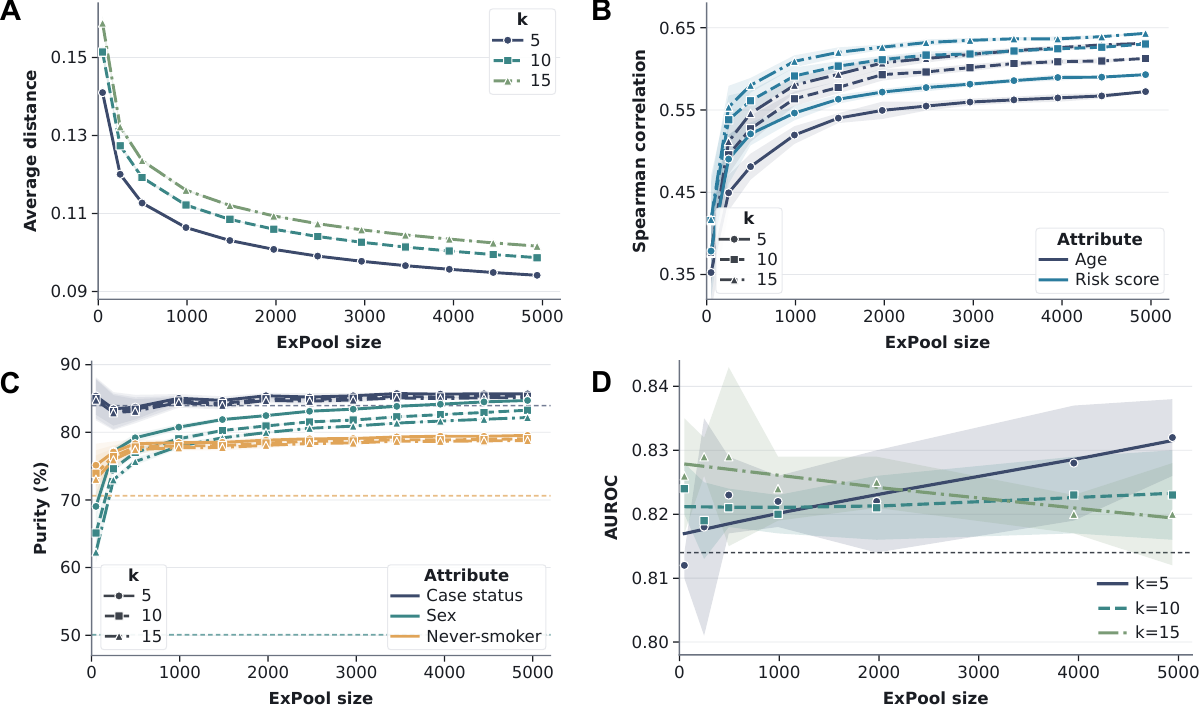}

\caption{\textbf{Evolution of the ExPool.} As ExPool size increases: \textbf{A,} average distance to retrieved neighbors decreases; \textbf{B,} index-neighbor Spearman correlations (age, risk score) increase; and \textbf{C,} retrieval purity (case status, sex, never-smoker) improves. Dashed lines in \textbf{C} indicate random retrieval baselines. \textbf{D,} AUROC trajectories for $k \in \{5, 10, 15\}$ retrieved patients (mean of 3 seeds). The dashed line denotes the baseline without ExPool (0.814).}
\vspace{-10pt}
\label{fig:evolving_expool}
\end{figure}

\paragraph{Retrieval quality improves monotonically with pool size.} As ExPool grows from <100 to 5,000 verified patients, average embedding distance between an index patient and its top-$k$ neighbors decreases (Figure \ref{fig:evolving_expool}A), Spearman correlations on age and predicted risk increase (Figure \ref{fig:evolving_expool}B), and purity by case status, sex, and smoking status rises well above random retrieval baselines (Figure \ref{fig:evolving_expool}C). The pool therefore retrieves progressively more clinically relevant neighbors as experience accumulates.

\paragraph{Optimal $k$ shifts from diversity to specificity.} Predictive AUROC (Figure \ref{fig:evolving_expool}D) reveals a tradeoff between retrieval diversity and specificity. When ExPool is small, a larger $k$ is best, potentially leveraging diversity to compensate for sparse coverage; once ExPool matures, a smaller $k$ wins, indicating that dense pools may prefer specificity-driven retrieval. Across all sizes, the few-shot framework remains above the zero-shot static Traj-CoA baseline.

\subsection{Evolutionary Dynamics of MARL}
\begin{figure}[h!]
\centering
\includegraphics[width=\linewidth]{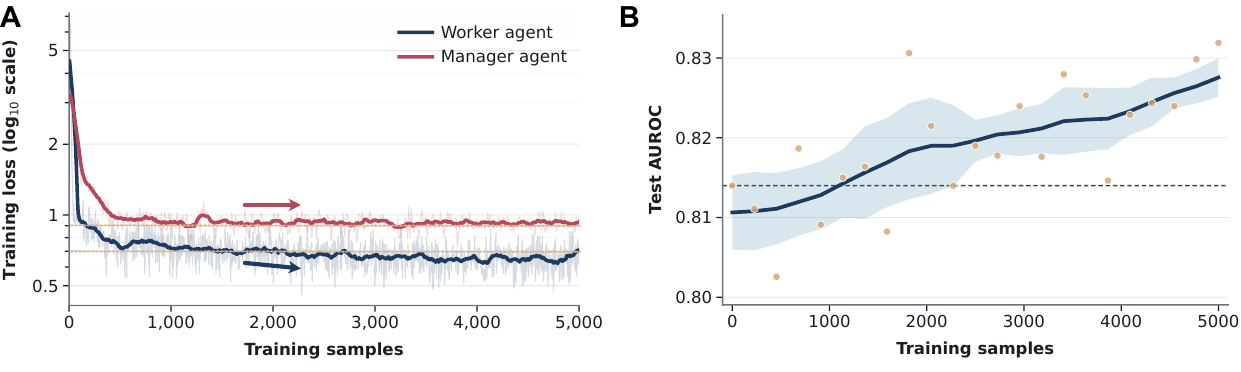}
\caption{
\textbf{Evolving MARL performance during agent training.}
\textbf{A,} Loss curves for the worker and manager agents across training iterations.
\textbf{B,} Model performance (mean of 3 seeds) as the number of training samples increases.
}
\vspace{-10pt}
\label{fig:evolving_rl}
\end{figure}

\paragraph{Decoupled training yields asymmetric learning curves.} Worker and manager agents both show sharp initial loss decreases followed by stabilization (Figure \ref{fig:evolving_rl}A). However, the manager loss converges quickly, while the worker loss continues to decline through 5,000 training samples. This is consistent with the intuition that manager summarization can be easier to internalize than fine-grained, sequential temporal reasoning and collaborative memory distillation across worker agents and the LTM.

\paragraph{Test AUROC scales with verified experience}
Test AUROC rises from about 0.81 to 0.83 as the MARL training pool grows to 5,000 samples (Figure \ref{fig:evolving_rl}B). The simultaneous decrease of worker loss and continued AUROC gains suggest that optimizing temporal reasoning may bridge the accumulation of verified experience and task performance.

\subsection{Mechanism Analysis}
\begin{figure}[h!]
    \centering
    \includegraphics[width=\linewidth]{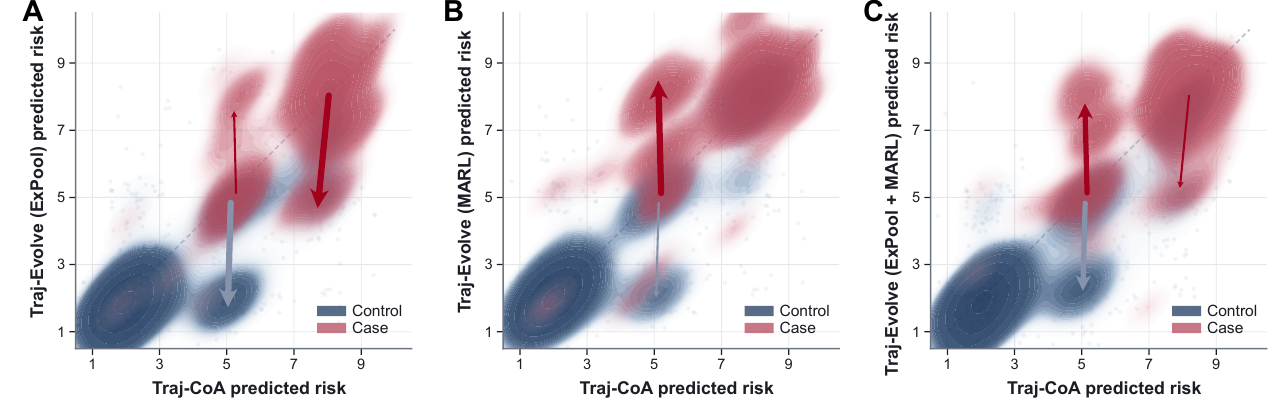}
    \caption{\textbf{Optimization properties of Traj-Evolve's self-evolving mechanisms.} Density scattered plots comparing the predicted risk scores of Traj-Evolve variants against the static Traj-CoA baseline. Arrows illustrate the strength of how Traj-Evolve changes the scores over Traj-CoA (wider means stronger). Scattered points are presented in a jittered way to facilitate visualization.}
    \vspace{-10pt}
    \label{fig:score}
\end{figure}

\paragraph{ExPool and MARL exert complementary effects on the score distribution.} Density plots of Traj-Evolve risk scores against the static Traj-CoA (Figure \ref{fig:score}) reveal distinct optimization patterns of the two evolving strategies. ExPool acts primarily on the negative class, pulling control scores downward (improving specificity) at the cost of mildly depressing some case scores. MARL acts primarily on ambiguous mid-range cases, lifting them toward high risk estimates (improving sensitivity). Their combination preserves both effects, producing the cleanest separation between cases and controls. This analysis is also consistent with the observed performance metrics in Table \ref{tab:1_year_performance} and Table \ref{tab:1_year_performance_nonsmoker}, where Traj-Evolve generally has a higher specificity with ExPool, a higher sensitivity with MARL, and a balanced sensitivity and specificity for the combined system.

\section{Conclusion}
We present Traj-Evolve, a self-evolving multi-agent framework for lung cancer early detection from longitudinal EHRs. By combining non-parametric few-shot retrieval (ExPool) with parametric reasoning internalization (MARL), Traj-Evolve emulates the continuous learning of experienced clinicians. Traj-Evolve establishes a new state-of-the-art for one-year lung cancer prediction and shows particular robustness in challenging subgroups like never-smokers. Furthermore, we demonstrate that ExPool and MARL are highly complementary, improving specificity and sensitivity, respectively. Future work will extend this framework to real-world streaming settings with richer reward signals and additional clinical tasks.

\section*{Limitations}
First, our evaluation was conducted within a single-institution retrospective case-control design. Prospective and multi-institutional validation will be essential to characterize calibration, generalizability across health systems and patient populations, and performance under genuine deployment prevalence.
Second, the self-evolving mechanisms depend on ground-truth diagnostic labels. In clinical practice, definitive labels for incident lung cancer may take months to years to materialize and are subject to verification noise, loss to follow-up, and ascertainment bias. The current framework also assumes that label-conditioned rejection sampling reliably identifies the optimal reasoning trace, yet a roll-out with the correct final prediction does not guarantee correct intermediate reasoning. Incorporating process-level reward signals \cite{zhangSurveyReinforcementLearning2025} may mitigate these issues.
Finally, this paper focused on one-year incident lung cancer prediction with a five-year look-back window. Extending Traj-Evolve to other outcomes, longer or shorter prediction horizons, and continuously updating patient trajectories in real-world streaming settings will be necessary to establish the framework as a general-purpose paradigm for self-evolving longitudinal EHR modeling.

\section*{Ethical Considerations}
We receive approval from the Institutional Review Board (IRB) to access the patient data. All code and data are stored and executed on HIPAA-compliant servers.

\section*{Acknowledgement}
This work was supported by the National Institutes of Health (NIH)—National Cancer Institute (Grant Nos. 1R01CA248422-01A1). Additional support was provided by the NIH under award number R35CA274442 to R.E. and the Rosalie and Harold Rea Brown Endowed Chair at Fred Hutch Cancer Center to R.E.

\bibliography{main}

\appendix

\section{Dataset Description}
\label{app:dataset}

This retrospective case-control study utilized an in-house longitudinal dataset to predict first primary lung cancer diagnoses within a one-year prediction period \cite{zhouRadTimelineTimelineSummarization2026}. The index date for prediction of both cases and controls was defined as the completion time of a qualifying chest-related radiology exam (chest CT, abdomen CT, or chest X-ray). To model patient trajectories, up to five years of EHR history prior to the index date was utilized. This multimodal EHR data encompasses both structured records (diagnosis, medication, lab, vital, and procedure) and unstructured text (clinical notes and radiology reports).

Eligible patients were over 40 years old. Cases were defined as individuals with a valid primary lung cancer diagnosis, excluding those with prior cancers. To ensure predictions informed early detection and excluded active diagnostic workups, the index exam for cases was required to be completed between two months and one year prior to diagnosis. Controls were defined as individuals with no cancer registry records of any cancer type; their index exams were selected to guarantee at least three years of cancer-free follow-up. To ensure longitudinal data, included patients should have an active clinical encounter history exceeding 120 days within the prior five years. Controls were matched to cases based on the index exam type and date (overlapping within a three-month window) using a 1:10 case-to-control matching schema. 

The matched cohort was randomly partitioned into mutually exclusive training (n=13,629) and validation (n=300) sets, with the remainder assigned to a held-out test set. To assess model performance and generalizability, we constructed two evaluation cohorts via independent random sampling from this test set: an overall cohort of 1000 patients (90 cases, 910 controls) encompassing all smoking statuses, and a never-smoker cohort of 835 patients (27 cases, 808 controls).

Baseline demographic, clinical, and lifestyle characteristics of the two test cohorts are summarized in Table \ref{tab:baseline}.
For the overall population, patient trajectories spanned a median of 4.5 years for cases and 3.7 years for controls, and contained tens to hundreds of dated entries drawn from diagnoses, procedures, laboratories, medications, vital signs, and free-text notes. 
The per-patient input was also exceptionally long, with median XML token counts exceeding 60,000 in the overall sample and 80,000 in the never-smoker cases. Time-aware chunking reduced the mean per-chunk token count to approximately 16,000, allowing Traj-Evolve to process the full trajectory effectively while preserving chronology.

\begin{table*}[htbp]
\centering
\footnotesize
\setlength{\tabcolsep}{4pt}
\renewcommand{\arraystretch}{1.0}
\caption{Baseline characteristics of cases and controls in the overall and never-smoker cohorts.}
\label{tab:baseline}
\begin{threeparttable}
\resizebox{\textwidth}{!}{%
\begin{tabular}{@{}lcccc@{}}
\toprule
 & \multicolumn{2}{c}{\textbf{Overall sample}} & \multicolumn{2}{c}{\textbf{Never-smoker sample}} \\
\cmidrule(lr){2-3} \cmidrule(lr){4-5}
\textbf{Characteristic} & Cases (n=90) & Controls (n=910) & Cases (n=27) & Controls (n=808) \\
\midrule
\multicolumn{5}{@{}l}{\textit{Demographics}} \\
\addlinespace[2pt]
Age at index date, years, median (IQR) & 68 (60--76) & 68 (59--78) & 74 (64--78) & 68 (58--80) \\
\addlinespace[2pt]
Sex, n (\%) & & & & \\
\quad Female & 49 (54.4) & 423 (46.5) & 24 (88.9) & 463 (57.3) \\
\quad Male   & 41 (45.6) & 487 (53.5) & 3 (11.1)  & 345 (42.7) \\
\addlinespace[2pt]
Race, n (\%) & & & & \\
\quad White & 74 (82.2) & 657 (72.2) & 21 (77.8) & 572 (70.8) \\
\quad Black or African American & 8 (8.9) & 102 (11.2) & 0 (0.0) & 80 (9.9) \\
\quad Asian & 5 (5.6) & 77 (8.5) & 6 (22.2) & 99 (12.3) \\
\quad Other & 1 (1.1) & 28 (3.1) & 0 (0.0) & 22 (2.7) \\
\quad Unknown & 2 (2.2) & 46 (5.1) & 0 (0.0) & 35 (4.3) \\
\addlinespace[2pt]
Ethnicity, n (\%) & & & & \\
\quad Hispanic or Latino & 4 (4.4) & 48 (5.3) & 4 (14.8) & 47 (5.8) \\
\quad Not Hispanic or Latino & 80 (88.9) & 758 (83.3) & 23 (85.2) & 722 (89.4) \\
\quad Unknown & 6 (6.7) & 104 (11.4) & 0 (0.0) & 39 (4.8) \\
\midrule
\multicolumn{5}{@{}l}{\textit{Clinical index and EHR history}} \\
\addlinespace[2pt]
Index year, median (IQR) & 2016 (2014--2018) & 2016 (2014--2018) & 2018 (2016--2018) & 2017 (2015--2018) \\
EHR look-back duration, years, median (IQR) & 4.5 (2.0--4.9) & 3.7 (1.1--4.9) & 4.2 (2.4--4.8) & 4.5 (2.8--4.9) \\
Time to event, years, median (IQR) & 0.5 (0.4--0.7) & --- & 0.5 (0.3--0.8) & --- \\
XML token count, median (IQR) & 61\,678 (26\,535--151\,282) & 51\,904 (18\,429--125\,871) & 84\,406 (46\,292--131\,768) & 75\,070 (35\,060--153\,634) \\
Distinct dated EHR entries, median (IQR) & 44 (19--113) & 39 (12--95) & 74 (30--102) & 58 (28--114) \\
\addlinespace[2pt]
Index exam modality, n (\%) & & & & \\
\quad Computed radiography (CR) & 45 (50.0) & 516 (56.7) & 15 (55.6) & 493 (61.0) \\
\quad Computed tomography (CT) & 45 (50.0) & 394 (43.3) & 12 (44.4) & 315 (39.0) \\
\addlinespace[2pt]
Index exam body site, n (\%) & & & & \\
\quad Chest & 59 (65.6) & 548 (60.2) & 18 (66.7) & 550 (68.1) \\
\quad Abdomen/pelvis & 8 (8.9) & 135 (14.8) & 5 (18.5) & 140 (17.3) \\
\quad Extremity & 5 (5.6) & 95 (10.4) & 2 (7.4) & 49 (6.1) \\
\quad Other & 0 (0.0) & 4 (0.4) & 0 (0.0) & 4 (0.5) \\
\quad Missing & 18 (20.0) & 128 (14.1) & 2 (7.4) & 65 (8.0) \\
\addlinespace[2pt]
Low-dose CT lung cancer screening, n (\%) & 1 (1.1) & 18 (2.0) & 0 (0.0) & 0 (0.0) \\
\midrule
\multicolumn{5}{@{}l}{\textit{Clinical text and structured data volume, median (IQR)}} \\
\addlinespace[2pt]
Epic notes & 10 (0--50) & 6 (0--28) & 32 (7--58) & 20 (8--45) \\
ORCA notes & 2 (0--12) & 4 (0--13) & 0 (0--10) & 2 (0--12) \\
Radiology reports & 10 (5--19) & 8 (3--14) & 9 (6--18) & 8 (3--15) \\
Diagnosis entries & 144 (53--340) & 118 (34--346) & 283 (115--377) & 202 (86--451) \\
Medication entries & 52 (9--200) & 59 (10--164) & 77 (37--173) & 72 (23--184) \\
Procedure entries & 118 (37--220) & 98 (34--221) & 100 (72--170) & 124 (52--272) \\
Laboratory entries & 167 (54--538) & 177 (49--484) & 293 (160--627) & 230 (93--577) \\
Vital-sign entries & 28 (0--118) & 18 (0--83) & 81 (22--172) & 61 (24--135) \\
\midrule
\multicolumn{5}{@{}l}{\textit{Lifestyle factors}} \\
\addlinespace[2pt]
Smoking status, n (\%) & & & & \\
\quad Current & 16 (17.8) & 82 (9.0) & 0 (0.0) & 0 (0.0) \\
\quad Former & 29 (32.2) & 180 (19.8) & 0 (0.0) & 0 (0.0) \\
\quad Never & 11 (12.2) & 289 (31.8) & 27 (100.0) & 808 (100.0) \\
\quad Passive exposure only & 0 (0.0) & 2 (0.2) & 0 (0.0) & 0 (0.0) \\
\quad Unknown & 34 (37.8) & 357 (39.2) & 0 (0.0) & 0 (0.0) \\
\addlinespace[2pt]
Alcohol use, n (\%) & & & & \\
\quad Current use & 16 (17.8) & 219 (24.1) & 10 (37.0) & 285 (35.3) \\
\quad No current use & 30 (33.3) & 260 (28.6) & 11 (40.7) & 426 (52.7) \\
\quad Unknown & 44 (48.9) & 431 (47.4) & 6 (22.2) & 97 (12.0) \\
\bottomrule
\end{tabular}%
}
\end{threeparttable}
\end{table*}

\section{Experimental Settings}
\subsection{Implementation Details}
\label{app:implementation}

We serve all LLMs using vLLM 0.19.1 \cite{kwonEfficientMemoryManagement2023} 
and implement MARL with QLoRA \cite{dettmersQLoRAEfficientFinetuning2023} 
via the \texttt{unsloth} library (v2026.2.1)\footnote{https://unsloth.ai/} on 4 GPUs. Both the worker and manager 
agents share the same training configuration: LoRA rank and $\alpha$ are both set 
to 32, with gradient checkpointing enabled to reduce memory overhead. We use a 
per-device batch size of 1 with gradient accumulation over 8 steps, yielding an 
effective batch size of 8 per device. Models are trained for a single epoch with 
a linear learning rate schedule, a peak learning rate of $2\times10^{-4}$, 5 warmup 
steps, and weight decay of 0.01. Optimization is performed using 8-bit AdamW. 
For LLM-as-a-judge evaluation, we use GPT-OSS-120B as the judge model.

\subsection{Baselines}
\label{app:baselines}

We compare against baselines from five categories: (i) a clinical risk model, LCRAT \citep{katki2016development}; (ii) supervised machine learning models, logistic regression and XGBoost \citep{chenXGBoostScalableTree2016}; (iii) sequential deep learning models, RETAIN \citep{choi2016retain} and PatientTM \citep{silvaModellingPatientTrajectories2022a}; (iv) a clinical BERT encoder, Clinical ModernBERT \citep{leeClinicalModernBERTEfficient2025}; and (v) LLM-based pipelines built on GPT-OSS-20B \citep{openaiGptoss120bGptoss20bModel2025} with or without long-context modeling strategies, including vanilla single-agent LLM, RAG, and Traj-CoA \citep{zeng2025trajcoa}. 

To ensure feasible and fair comparisons, input data modalities were tailored to each model's specific architectural requirements. Standard ML models utilized structured medical codes, whereas PatientTM incorporated both codes and unstructured clinical texts (i.e., clinical notes and radiology reports). For BERT- and LLM-based architectures, we employed the unified XML-formatted EHR representation that preserved multimodal information, including codes, free text, and numerical laboratory values. For long sequences exceeding model context limits, left-truncation was applied to prioritize the most recent clinical records. 

Supervised baselines, Clinical ModernBERT, and the MARL-optimized Traj-Evolve were trained on the same training set. In contrast, the ExPool variant of Traj-Evolve operated in a few-shot manner, dynamically retrieving ``patients-like-me'' from the ExPool constructed using the training set, while zero-shot baselines operated without access to the training set. Optimal hyperparameter configurations for all trained baselines were determined using the validation set. During evaluation, the threshold that maximizes Youden's J-index \cite{ruoppYoudenIndexOptimal2008} was used.

\section{Case Study}
To qualitatively demonstrate the nuanced clinical reasoning capabilities enabled by the ExPool, we examined the diagnostic trajectory of a 50-year-old Asian female never-smoker with a history of occupational asbestos exposure. The index patient presented with a normal pulmonary function test (PFT) and cough in 2015, formally documented asbestos exposure in 2016, and the discovery of a 16 mm non-calcified left upper lobe (LUL) nodule in 2018, with normal PFT and absence of pulmonary symptoms. She was subsequently diagnosed with lung cancer.

Traj-Evolve used information from two of the retrieved 10 patients from ExPool. Patient 8 was a 55-year-old woman who shared documented asbestos exposure and progressive pulmonary fibrosis, but lacked a discrete mass and remained cancer-free after 3 years. Patient 10 was a 57-year-old Asian female never-smoker who presented with an 18 mm LUL mass and was confirmed to have lung cancer within 6 months.

In its generated reasoning rationale, Traj-Evolve explicitly estimated the index patient's risk by comparing her longitudinal trajectory to these retrieved cohorts. The manager agent identified that the index patient ``sits biologically between these extremes'', correctly weighing the mitigating factor of normal pulmonary function against the high-risk modifiers of progressive environmental carcinogen exposure and a suspicious nodule. The system assigned a risk score of 7/10. This case exemplifies how the incorporation of verified clinical experience enables Traj-Evolve to execute robust, comparative clinical judgment in atypical and highly challenging clinical presentations.

\begin{figure*}[htbp!]
\centering
\includegraphics[width=\linewidth]{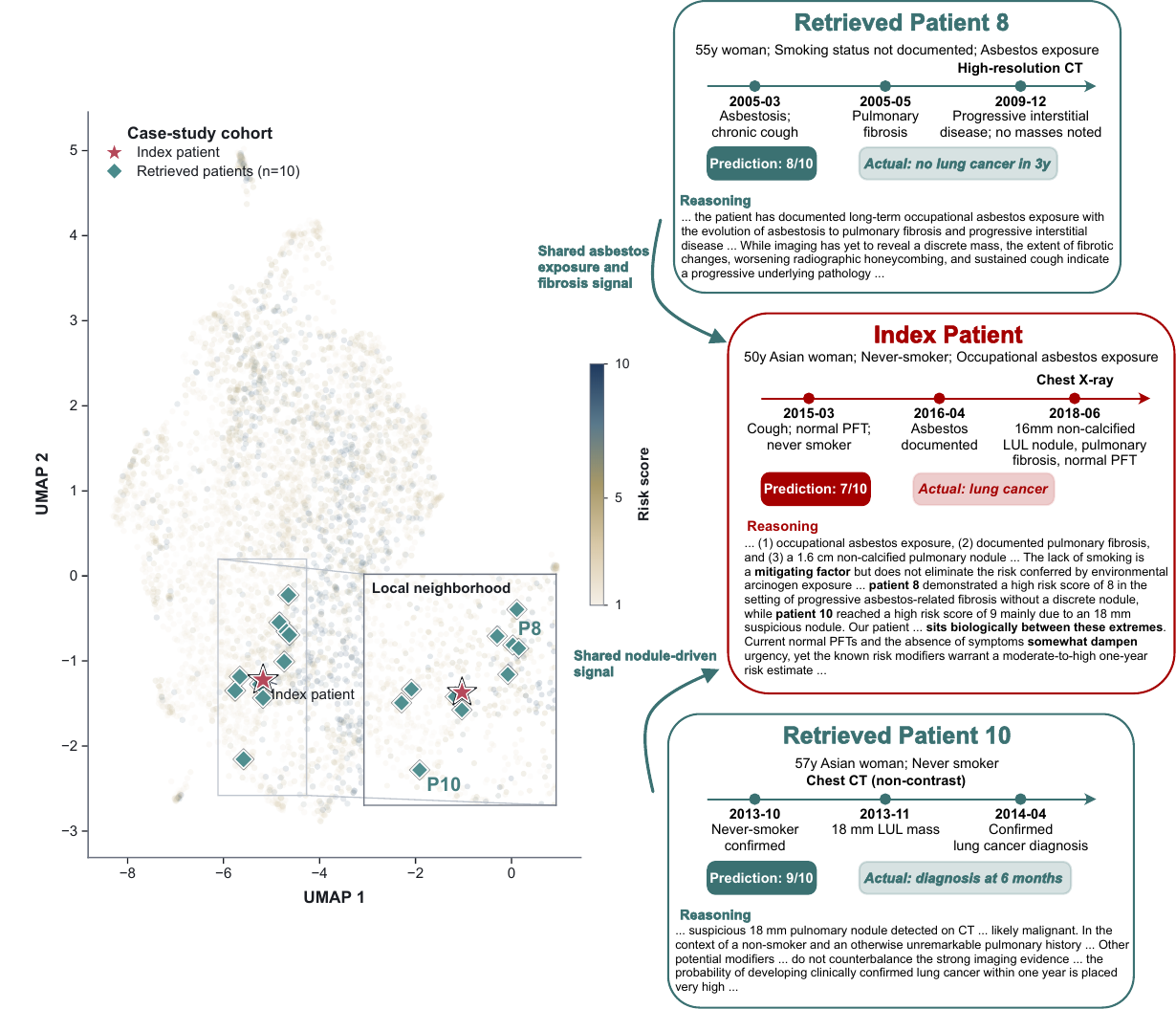}
\caption{
Case study demonstrating Traj-Evolve (ExPool+MARL)'s reasoning for a never-smoker patient. A UMAP projection maps the local semantic neighborhood of the index patient (star) alongside retrieved historical cases (diamonds). The right panel demonstrates how Traj-Evolve balances the shared signals from retrieved patients and the index patient's own characteristics.
}
\label{fig:case_study}
\end{figure*}

\section{Prompts}
\label{app:prompts}

We follow the same prompt template in Traj-CoA \cite{zeng2025trajcoa} for Traj-Evolve without ExPool. With ExPool, the manager agent's system prompt and user prompt are shown in Table \ref{tab:prompt_aggregation_agent_system_with_pool} and \ref{tab:prompt_aggregation_agent_user_with_pool}.

\begin{table*}[ht]
\centering
\caption{System prompt for the manager agent with ExPool (Patients-Like-Me retrieval).}
\small
\begin{tabularx}{\textwidth}{X}
\toprule
\underline{\textbf{\textsc{Manager Agent System Prompt with ExPool}}} \\
\midrule
You are a senior clinical AI expert specializing in longitudinal lung cancer risk analysis. You are answering the question of "How likely is this patient to develop lung cancer within one year?" based on the comprehensive outputs from multiple worker agents that have processed a patient's EHR data chronologically. \\
\\
\textbf{Task:} Synthesize the outputs from the last worker agent and the universal memory of all lung cancer related events to provide a final, comprehensive lung cancer risk assessment and a narrative of the patient's risk evolution. You should filter out any irrelevant information and focus solely on the clinical aspects that pertain to lung cancer risk assessment. \\
\\
\textbf{Input:} \\
- \texttt{final\_worker\_outputs}: A JSON object, which is the output from the last worker agent that has processed a patient's EHR data chronologically. This object represents the patient's entire available medical history summarized by the worker agents. \\
- \texttt{universal\_memory\_events}: A list of all lung cancer related events from the universal memory, providing complete historical context across all processed chunks. \\
- \texttt{patients\_like\_me}: A list of similar patients retrieved from historical records. Each entry contains a clinical summary, the model's previous prediction, and the actual diagnosis outcome. \textbf{Important:} These reference cases are provided to help you learn from past experience—both successes and failures—but they may or may not be relevant to the current patient. \\
\\
\textbf{Instructions:} \\
1.  \textbf{Synthesize Temporal Trends:} Review the sequence of outputs and the complete universal memory. Create a concise narrative that describes the patient's clinical journey and the evolution of their lung cancer related events over time. Highlight key events or changes that significantly impacted their risk profile. \\
2.  \textbf{Final Lung Cancer Related Events Assessment:} Consolidate all identified lung cancer related events from the universal memory and worker outputs into a final, comprehensive list. Ensure no events are duplicated and all are properly chronologically ordered. \\
3.  \textbf{Learn from Similar Patients (Patients-Like-Me):} Critically evaluate each retrieved similar patient case: \\
    \quad - \textbf{Assess Relevance:} Determine whether each case is truly comparable to the current patient based on demographics, clinical presentation, risk factors, and disease trajectory. Some cases may appear similar but differ in critical ways. \\
    \quad - \textbf{Learn from Correct Predictions:} When the model's prediction matched the actual outcome, identify what clinical reasoning led to the correct assessment. \\
    \quad - \textbf{Learn from Incorrect Predictions:} When the model's prediction was wrong, analyze what was missed or overweighted. Use these as cautionary examples to avoid similar errors. \\
    \quad - \textbf{Selective Reference:} You are NOT required to use all or any of these cases. Only reference those that genuinely inform your reasoning. Explicitly state if none are sufficiently relevant. \\
    \quad - \textbf{Caution:} Avoid just averaging the risk assessments of the similar patients. Instead, use the information from the similar patients to inform your reasoning. \\
4.  \textbf{Assess Final Lung Cancer Risk:} Provide a final lung cancer risk assessment, from 1 to 10, where 1 is the lowest risk and 10 is the highest risk. \\
5.  \textbf{Provide Comprehensive Reasoning:} Justify your final risk assessment by explaining how the interplay of all lung cancer related events from the universal memory and their temporal evolution contributes to the patient's overall risk. If you referenced any similar patient cases, briefly explain how they informed your decision. This should be your most detailed and conclusive reasoning. \\
\\
\textbf{Output Format:} \\
Your output must be a single, easily parsable JSON object with the following keys: \\
- \texttt{risk\_evolution\_summary}: A string containing the narrative of the patient's clinical journey and risk evolution. \\
- \texttt{final\_lung\_cancer\_related\_events}: A list of strings containing all unique, consolidated lung cancer related events from the universal memory. \\
- \texttt{final\_risk\_assessment}: A JSON object for the final risk level for lung cancer diagnosis within 1 year (1 to 10, where 1 is the lowest risk and 10 is the highest risk). \\
    \quad - \texttt{risk\_level}: An integer from 1 to 10, where 1 is the lowest risk and 10 is the highest risk. \\
    \quad - \texttt{reasoning}: A string providing a comprehensive justification for the final risk assessment, including any insights drawn from similar patient cases. \\
\\
ONLY output the JSON object without any additional text or formatting. Ensure that the JSON is valid and can be parsed easily. \\
\bottomrule
\end{tabularx}
\label{tab:prompt_aggregation_agent_system_with_pool}
\end{table*}

\begin{table*}[ht]
\centering
\caption{User prompt for the manager agent with ExPool.}
\small
\begin{tabularx}{\textwidth}{X}
\toprule
\underline{\textbf{\textsc{Manager Agent User Prompt with ExPool}}} \\
\midrule
All Worker Agent Outputs: \\
\texttt{<final\_worker\_outputs>} \\
\{\texttt{final\_worker\_outputs}\} \\
\texttt{</final\_worker\_outputs>} \\
\\
Universal Memory Events (All Events): \\
\texttt{<universal\_memory\_events>} \\
\{\texttt{universal\_memory\_events}\} \\
\texttt{</universal\_memory\_events>} \\
\\
Similar Patients from Historical Records (Patients-Like-Me): \\
\texttt{<patients\_like\_me>} \\
\{\texttt{patients\_like\_me}\} \\
\texttt{</patients\_like\_me>} \\
\\
Please provide the final risk assessment and narrative summary in JSON format. \\
\bottomrule
\end{tabularx}
\label{tab:prompt_aggregation_agent_user_with_pool}
\end{table*}

\end{document}